# Enhancing Track Management Systems with Vehicle-To-Vehicle Enabled Sensor Fusion

Thomas Billington
*Department of Computer Science*
*Virginia Polytechnic Institute and State University*
Blacksburg, VA, USA
tommybillington@vt.edu

Ansh Gwash
*Department of Computer Science*
*Virginia Polytechnic Institute and State University*
Blacksburg, VA, USA
anshgwash@vt.edu

Aadi Kothari
*Department of Electrical and Computer Engineering*
*Virginia Polytechnic Institute and State University*
Blacksburg, VA, USA
aadikothari@vt.edu

Lucas Izquierdo
*Department of Electrical and Computer Engineering*
*Virginia Polytechnic Institute and State University*
Blacksburg, VA, USA
lucasizq@vt.edu

Timothy Talty
*Department of Electrical and Computer Engineering*
*Virginia Polytechnic Institute and State University*
Blacksburg, VA, USA
ttalty@vt.edu

*Abstract—* **In the rapidly advancing landscape of connected and automated vehicles (CAV), the integration of Vehicle-to-Everything (V2X) communication in traditional fusion systems presents a promising avenue for enhancing vehicle perception. Addressing current limitations with vehicle sensing, this paper proposes a novel Vehicle-to-Vehicle (V2V) enabled track management system that leverages the synergy between V2V signals and detections from radar and camera sensors. The core innovation lies in the creation of independent priority track lists, consisting of fused detections validated through V2V communication. This approach enables more flexible and resilient thresholds for track management, particularly in scenarios with numerous occlusions where the tracked objects move outside the field of view of the perception sensors. The proposed system considers the implications of falsification of V2X signals which is combated through an initial vehicle identification process using detection from perception sensors. Presented are the fusion algorithm, simulated environments, and validation mechanisms. Experimental results demonstrate the improved accuracy and robustness of the proposed system in common driving scenarios, highlighting its potential to advance the reliability and efficiency of autonomous vehicles.**

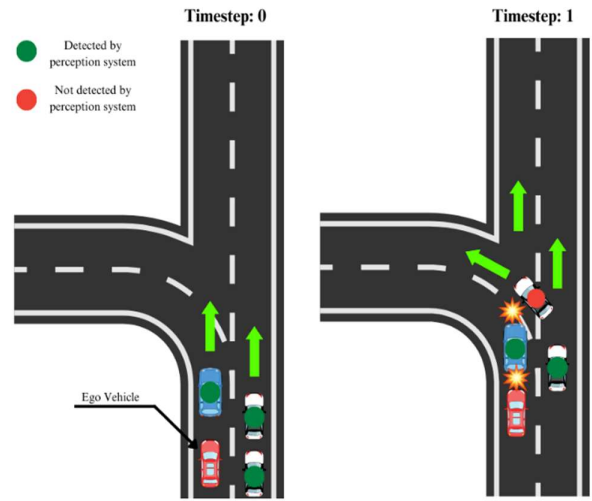

*Figure 1: Traditional sensor fusion failure scenario*

## I. Introduction

The emergence of connected vehicles has transformed the transportation industry, introducing advanced safety and increased traffic flow optimization. Central to the realization of these benefits is the development of advanced perception systems that can accurately detect, track, and predict the movement of surrounding vehicles and objects. Traditionally, sensor fusion systems have relied heavily on perception sensors including long-range radar, short-range radar, and cameras to survey the ego vehicle's environment. The limitations of these sensors' utility, especially in environments with significant occlusions, pose challenges to the reliability of current track management systems. To address these issues, V2V communication can be leveraged with local sensor fusion to enhance object tracking with the aim to create a more comprehensive representation of the surrounding environment. Utilization of validated track lists that have been formed from fusing detections of all on-vehicle perception sensors and confirmed through positioning data from Basic Safety Messages (BSM), increases the availability of object tracking through the extension of the track-management systems' effective range beyond the limitations of individual sensors. This is inherently important in Advanced Driver Assistance Systems (ADAS) such as Autonomous Intersection Navigation (AIN) and collision avoidance where sensor dropouts are detrimental to the resilience of the system as seen in Figure 1. Effectively and accurately perceiving the entire road, regardless of occlusions, is an invaluable asset and allows the path planning systems the ability to make reliable and safe decisions.

Striving for improvements in vehicle perception systems for autonomous vehicles is not only important for technological advancement but also for safety needs. A V2V-enabled track management system, as proposed, has the potential to significantly reduce vehicle crashes, which are a major cause of death or injury worldwide. By minimizing the chance for human error, especially in complex traffic scenarios, connected vehicles can provide a more efficient, robust, and safe experience when compared with traditional vehicles. While collision avoidance strategies attempt to fix this problem and have become commonplace in the industry, they only tackle emergency situations. The proposed V2V system expands on these conventional strategies by providing predictive capabilities that anticipate potential hazards before they become imminent threats. For example, at complex intersections or high-traffic areas, such as busy urban environments, the system could detect the positions of hidden vehicles that are occluded.

Admittedly, the incorporation of V2V into sensor fusion has introduced safety-critical scenarios in which proposed collaborative fusion systems allow equal perception between perception sensors and V2V signals. Dangers of self-telemetry manipulation including injection and GPS spoofing attacks [1] invite the possibility to undermine the availability of the ADAS feature due to the nature of relying on V2V as a significant contributor within the fusion system. Prior to confirmation of the signal's validity, substantial risk is taken in terms of data integrity and the potential for compromised safety in system functionalities. This issue is overcome by assigning the V2V system the goal of validating and continuously tracking vehicle position data following the confirmation of the vehicle's existence, allowing for a restrictive approach to collaborative perception. This ensures safety and prevents potential tampering, which has created numerous inquiries into how reliable current V2V systems can truly be.

## II. LITERATURE REVIEW

Trends in academia point to the idea that a vehicle perception stack benefits from V2V communication through BSMs, showing the lack of accurate positioning and trajectory of the host and other remote vehicles from perception sensors. Olenscki et al.[2] support this concept through a novel algorithm using V2V BSMs for the implementation of Intersection Movement Assist (IMA) and Lane Change Assist (LCA). Experiments are run on the CARLA simulator, reducing costs while ensuring faster development. The results are highly suggestive of a V2V BSM approach towards perception. V2V augments the efficacy of the system by mitigating sensor perception obstructions. Despite the advantages, using it solely over other sensors can be quite difficult, due to lack of scaling.

Threats on integrity, as detailed by Alnasser et al. [1], are an additional primary cause for concern in systems where V2V is present in their perception stack. The wireless communication standard utilized in V2V, IEEE802.11p, is susceptible to internal attacks where injected messages are sent over the network. This creates the opportunity for attackers to create potentially dangerous road situations that greatly compromise driver safety. Overcoming difficulties in the lack of uniform use of V2V technology on the road as well as possible falsification can be accomplished by utilizing the positioning messages as supplementary information rather than for use in primary perception.

Another work that enhances perception of perception sensors with V2V BSMs is Tian et al [3]. They propose an innovative method by improving the accuracy of data association with their proposed method. A V2V Dedicated Short Range Communication (DSRC) setup is fused with a radar sensor. The noisy measurements of radar are complemented by the GPS and identity (positioning, heading, velocity) information from the BSMs. With increasing number of targets, the computation for Joint Probabilistic Data Association (JPDA) grows exponentially potentially causing data overlap. The effectiveness of Nearest Neighbor (NN) methods also decreases in such cases. Their work identifies a multi-measurement process with a Kalman Filter (KF) to process and fuse tracks. Additionally, they also use RTK-GPS over a traditional GPS to enhance their performance. However, the work is conducted on an older pipeline, DSRC, and it has been freed up to make research progress on Cellular V2X (C-V2X), which is the technology aimed for use in our systems. Additionally, while their vehicle platform has other sensors like LiDAR and cameras, they do not utilize them for fusion. This is another shortcoming we aim to address by using various radars and a camera along with BSM information.

Recent developments in probabilistic data fusion frameworks demonstrate the potential for combining data from local sensors and V2V units to ensure extended perception of driving environments. Utilizing similar approaches is valuable in the handling of inherent uncertainties associated with sensor data [4]. Although, alleviating false measurements collected from perception sensors and V2V units requires additional synchronization. This allows for more robust track management through the efficient employment of coordinate transformation, filtering, and buffering. Data association in the case of autonomous vehicles is a challenging task due to ambiguity, noise, and occlusions in perception. More often than not, sensors will collect data that could correspond to multiple objects or tracks, especially in environments with significant data overlap. In these cases, using a Mahalanobis distance-based data association algorithm between data information collected from traditional sensors and V2V units allows for more reliable results, especially in potential collision scenarios [5]. The compatibility with the extended Kalman Filter and its ability to provide information on object status and their uncertainties, is also a factor when deciding on distance calculation methods for data association. Although the Host Vehicle (HV) outlined in the experimental framework only uses a camera and V2V-enabled On-Board Unit (OBU), it is easily modifiable to allow for other sensors.

## III. METHODOLOGIES

### A. Local Fusion Processes

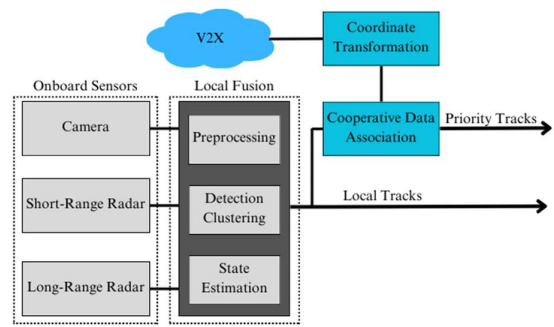

*Figure 2: High-level overview of V2V integrated fusion*

The proposed system seen in Figure 2 is designed to streamline sensor fusion and allow for flexible incorporation of V2V and perception sensors including long-range radar, short-range radar, and cameras. The detections received from these devices serve as input on which preprocessing steps are taken where sensor data is subjected to filtering and noise reduction techniques to enhance signal quality. For radar, this includes the application of thresholding algorithms to eliminate weak returns, while for camera data, image processing techniques including histogram equalization and edge detection are employed to improve feature visibility. Density-Based Spatial Clustering of Applications with Noise (DBSCAN) is well-suited for the clustering of closely localized radar detections and is ideal for dealing with the inherent variability of sensor data. Camera imaging is analyzed through computer vision techniques including the use of a convolutional neural network, You Only Look Once version 4 (YOLOv4), for object detection and classification. The processed detections from both camera and radar are then concatenated into a single list to prepare for future steps. State estimation [6] is performed through the usage of an Extended Kalman Filter to estimate the position, velocity, and dynamic properties of each individually tracked object. The culmination of the local fusion process results in the creation of a fully fused object track list consisting of reconstructed objects built from the various detections of vehicles within the local system's perceivable range. This local sensor track list serves as a baseline to later employ data association techniques between itself and the V2V BSM location information. This provides the decisioning system the ability to rely on a local list of detections to make decisions in vehicle movement, should there be a failure in the V2V-enabled system's ability to accurately perceive the environment.

*B. V2V Coordinate Transformation*

Basic Safety Messages are crucial for sharing real-time information about a vehicle's state including positioning, speed, and heading data. To ensure that this information is accurately interpreted by all receiving entities, coordinate transformation is necessary [7]. This system converts vehicle localization data from the global coordinate system, World Geodetic System 1984 (WGS-84), to a cartesian coordinate system [8] that ensures future cohesion with the onboard sensor fusion track list object positions and the orientation of the ego vehicle. Global coordinates are given in terms of latitude ($\phi$), longitude($\lambda$), and altitude($h$) above the reference ellipsoid. These are converted to Earth-Centered, Earth Fixed (ECEF) Cartesian coordinates (X, Y, Z) using the following formulas:

$$X = (N(\phi) + h)\cos(\phi)\cos(\lambda)$$
$$Y = (N(\phi) + h)\cos(\phi)\sin(\lambda)$$
$$Z = (\frac{b^2}{a^2} N(\phi) + h)\sin(\phi)$$

Where $a$ is the semi-major axis of the reference ellipsoid, $b$ is the semi-minor axis, and $N(\phi)$ is the radius of curvature in the prime vertical where:

$$N(\phi) = \frac{a^2}{\sqrt{a^2\cos^2(\phi) + b^2\sin^2(\phi)}}$$

The global ECEF coordinates (X, Y, Z) can be translated to the local ECEF frame by subtracting the coordinates of the local origin ($X_0$, $Y_0$, $Z_0$):

$$X_{local} = X - X_0$$
$$Y_{local} = Y - Y_0$$
$$Z_{local} = Z - Z_0$$

This effectively shifts the reference frame, so the local origin is the new origin (0,0,0) of the local ECEF coordinate system. Aligning the local ECEF coordinates with local axes can be accomplished by applying a rotation matrix derived from the latitude and longitude of the origin:

$$R = R_x(\phi_0) \cdot R_y(\lambda_0) \cdot R_z(\gamma)$$

Where $R_x$, $R_y$, and $R_z$ are the rotation matrices around the x, y, and z axes and $\gamma$ is an optional rotation angle to align with a specific local direction.

*C. Data Fusion and Track Validation*

Data association is accomplished in Algorithm 1 by grouping objects based on Mahalanobis distance [5] with a desired confidence interval threshold $\epsilon$. A new priority track list pair is returned with the validated tracks containing objects that have been corroborated with both the perception sensors and the V2V BSMs. Prioritization is crucial to the reliability of the autonomous system and ensures that the associated track list is free of inaccurate detections and possible falsified inputs from malicious V2V devices. Given this verified track list created from the complete cooperative sensor fusion process, ADAS features can now leverage the priority tracks for vehicles that are V2V-enabled as well as utilize the local tracks as a fallback for vehicles that do not carry V2V OBUs.

---

**Algorithm 1:** Association of Tracks

**Input:**
$S = \{(X_s, Y_s, Z_s) | s \in \text{Sensor tracks}\}$: Set of sensor tracks
$V = \{(X_v, Y_v, Z_v) | v \in \text{V2V tracks}\}$: Set of V2V tracks
$C$ = Covariance matric of the measurement error

**Output:**
$A = \{(s, v) | s \in S, v \in V\}$: Set of associated tracks

**Initialize:**
$\epsilon$ = Threshold value (desired confidence interval)
$A = []$
$z_v = (X_v, Y_v, Z_v)^T$
$z_s = (X_s, Y_s, Z_s)^T$
**for** $s \in S$:
    **for** $v \in V$:
        $D_M = \sqrt{(z_v - z_s)^T C^{-1} (z_v - z_s)}$
        **if** $D_M \leq \epsilon$:
            **add** pair $(s, v)$ to $A$
        **end**
    **end**
**end**
**return** $A$

---

*D. Simulation Environment*

There exists a range of both commercial and non-commercial toolchains that can be used to develop autonomous driving software, each with their advantages and disadvantages. MathWorks tools were selected for this work because of their rich toolboxes, visualization and analysis features, and high customizability. Specifically, the Automated Driving Toolbox was used to create the test scenarios, simulate radar and vision sensors, and extract actor ground truth information, providing a robust foundation for detailed analysis. To test the restricted approach to collaborative perception, a comprehensive design and simulation plan was developed. Complex driving scenarios including urban and highway environments were constructed using MathWorks Driving Scenario Designer and visualized with RoadRunner [9]. The implementation of the core perception system was carried out in MATLAB and Simulink including local sensor fusion, BSM signal formulation and processing, and data association. A key testing scenario designed for benchmarking of V2V-enabled sensor fusion in AIN was a four-way intersection [10] with an unprotected left turn as seen in Figure 3. This scenario involved multiple vehicles equipped with V2V capabilities and created a situation where the perception of the safety-critical vehicle was first established by the camera and radar sensors but later lost due to occlusions and removal from the perception track list. Through the incorporation of the BSM signals, the duration and accuracy of the track being held in the priority list could be monitored throughout the simulation to ensure safe and accurate perception. The scenario was run and comparably tested between local sensor fusion and cooperative fusion.

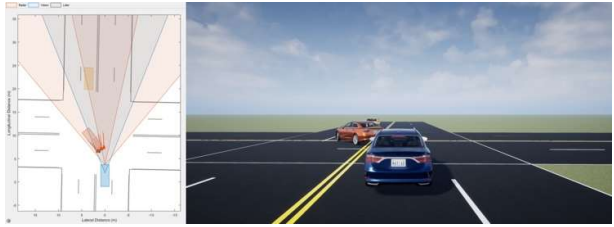

*Figure 3: 4-way intersection scenario modeled in MATLAB*

IV. EXPERIMENTAL RESULTS

*A. Tracking Metrics*

When comparing the tracking ability of the V2V-enabled fusion approach to that of the traditional sensor fusion system, several Key Performance Indicators (KPI) were set to ensure effective benchmarking. Generalized Optimal Sub-pattern Assignment (GOSPA), an extension of the Optimal Sub-pattern Assignment (OSPA), was used as a more specific metric [11], providing a comprehensive evaluation suited to complex multi-object tracking systems, capturing both the accuracy of object localization and the correctness of object enumeration [12]. GOSPA serves to calculate the distance between the estimated set of tracks and the true set of tracking considering localization, missed detections, and false tracks. It leverages these factors and allows for a validation metric that balances between penalizing localization errors versus missed and false tracks. Additional components of GOSPA include missed target error, false track error and switching error. These metrics were all used to evaluate the accuracy of the three different track management systems.

*B. Generalized Optimal Sub-pattern Metric Results*

The GOSPA metric is known for its robustness in quantifying the accuracy of object localization. Throughout the simulation experiment, the GOSPA metric was employed to measure the accuracy of three distinct setups: independent V2V communication tracks, local sensor fusion tracks from radar and camera detections, as well as the V2V-enabled sensor fusion tracks. The results from this experiment, as seen in Figure 4, revealed higher GOSPA metric values for the local sensor fusion tracks over the time interval with a mean value of 56.12, suggesting a higher rate of localization and cardinality error. In contrast, the V2V tracks demonstrated the lowest GOSPA metric values with a mean value of 7.517, highlighting its increased performance attributed to the additional data and perspective that the technology provides. The priority V2V-enabled sensor fusion tracks, which integrate both local sensor data and V2V communications, achieved relatively intermediary GOSPA metric values with a mean value of 48.62. This outcome suggests a synergistic effect where the integration of independent V2V BSM information helps mitigate the localization errors inherent in standalone local sensor fusion systems.

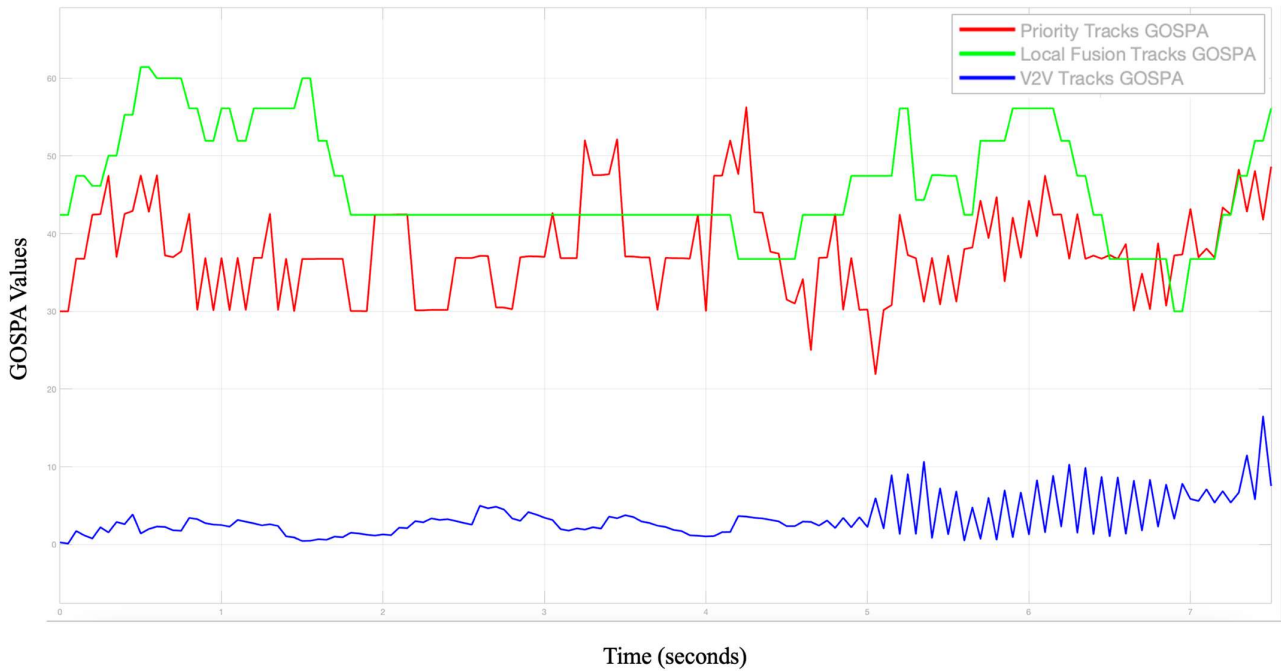

*Figure 4: GOSPA metric values over simulation interval*

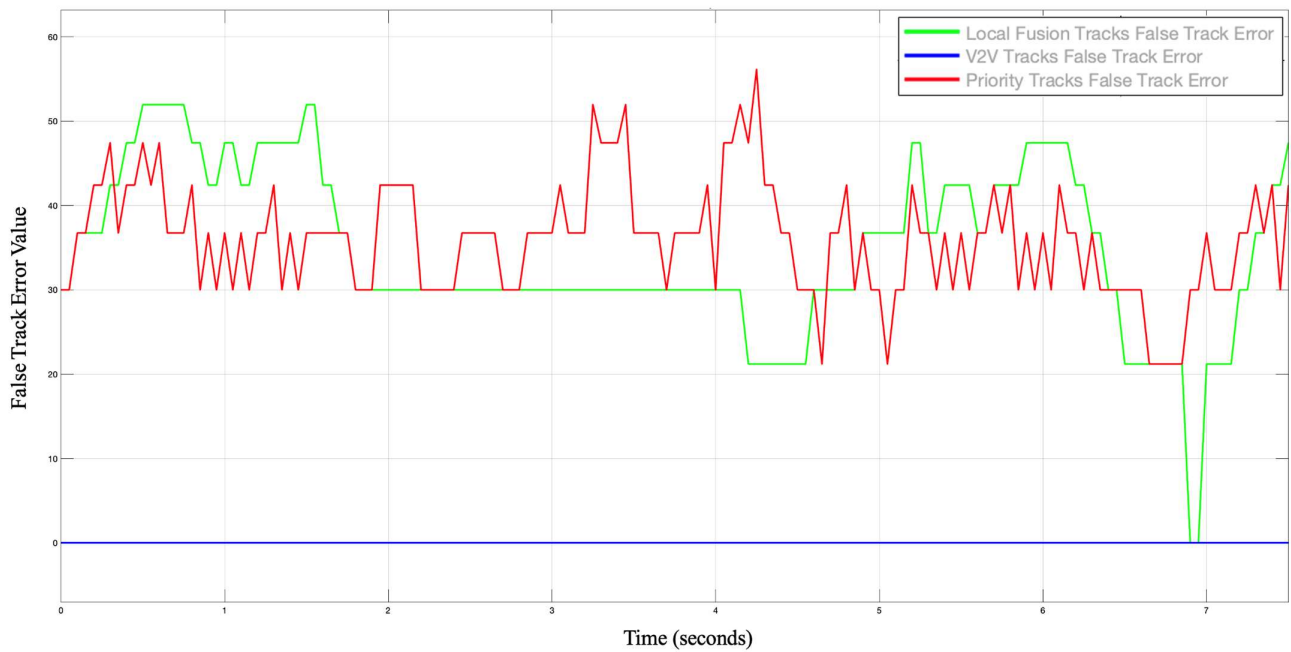

*Figure 5: False track error metric values over simulation interval*

### C. Missed Target Error Results

Determining Missed Target Error allows us to detect when object or entity detections fail to be tracked by the fusion system. This is important specifically in the intersection scenario we are testing, due to the occlusions that occur throughout the simulation. This scenario was designed to test the system's ability to continuously track vehicles outside the direct field of vision of the local sensors. The average Missed Target Errors over the simulation for local sensor fusion tracks, V2V tracks, and priority tracks were 30.0, 0.0, and 21.2 respectively. This disparity in Missed Target Errors suggests that V2V technology plays a crucial role in enhancing detection capabilities, particularly in situations where sensors alone might struggle due to occlusions. The significantly lower score that was noted for the V2V tracks highlights its effectiveness in maintaining continuous awareness of the driving environment and helps compensate for the visibility gaps in perception that are noted with the local sensor fusion system.

### D. False Track Error Results

False Track Error is a concern in sensor fusion systems, where the reliability of detecting objects is a significant priority. False tracks may prompt inappropriate actions, such as unnecessary braking or swerving, which compromises the safety of road users and disrupts traffic flow. Minimizing False Track Error is important to ensure safe reliance on the real-time tracking and decision-making process. Throughout the experiment, local sensor fusion tracks, V2V tracks, and priority tracks noted average False Track Error values of 47.43, 0.0, and 42.43 respectively. Results show a moderate decrease in False Track Errors when comparing the local fusion tracks and the priority tracks. The reduction in False Track Error can likely be attributed to the enhanced track management system. By leveraging data shared between vehicles through V2V signals, the tracking system effectively reduces the number of false tracks and improves the fidelity of the system. Additionally, the system can verify the continuity of objects across data frames, reducing the possibility of false tracks caused by erratic sensor noise or data losses.

### E. Switching Error Results

Maintaining consistent track identities is important throughout all stages of the sensor fusion process. Reducing Switching Errors allows for a more accurate identification of situations where the tracking system incorrectly swaps the identities of two or more targets, leading to inaccuracies in the system's output. All three tracking systems identified an average Switching Error of 0.0 across the time interval. The absence of Switching Errors across the multiple systems indicates a high level of accuracy within the simulation environment. However, it is important to note that in the systems may exhibit over-optimal performance as compared to Hardware-in-the-Loop (HIL) testing.

## V. Future Directions

To further evaluate and refine the effectiveness of our proposed V2X-enabled sensor fusion system, additional software testing as well as in-vehicle testing will be conducted. By emphasizing these methodologies, we can better address emerging challenges and explore the fusion algorithm's potential within a fully integrated vehicle system.

### A. Model-in-the-Loop (MIL) Testing

MIL testing involving Software Defined Radios, will allow us to include wireless communication protocols and external variables into the testing environment. This step is particularly important to verify the robustness of our algorithm under different noise-adverse conditions so that the performance can be improved before it is physically tested.

### B. Hardware-in-the-Loop (HIL) Testing

Finally, HIL testing brings in hardware to observe real-time interactions with our enhanced track management system. This is the final check to ensure seamless integration of the V2X radios into the vehicle perception system.

These testing protocols are integral parts of our development cycle to provide confidence in the performance and reliability of our proposed system. This comprehensive approach ensures a safe transition from simulation to real-world testing with validations at every stage.

## VI. Conclusions

In this study, we developed a V2V-enabled track management system that employs restrictive utilization of BSM localization information in vehicle perception. Our system introduces a novel integration of camera and radar track lists with V2V BSM data, overcoming difficulties noted in previous literature that treated these elements separately [3, 5]. The limitations of relying entirely on perception sensors can be detrimental to the reliability of the system in situations where occlusions are prevalent. Utilizing V2V information not only extends the operational range of the track management systems but also improves the resilience of safety-critical features such as AIN.

Our approach allows for the reliable fusion of detections from perception sensors and V2V units, seizing the full potential of range-extension benefits of V2V while minimizing risks associated with false detection. The reliance on V2V communication introduces challenges such as vulnerabilities to GPS spoofing and data injection. The integration of priority track lists enhances the ego vehicle's perception by combining data from local and V2V tracks. We also introduce a critical unprotected left-turn scenario to demonstrate the effectiveness of an enhanced V2V based sensor fusion.

The overall results of this experiment underscore the benefits of integrating V2V technology within traditional sensor fusion systems. Across all of our outlined metrics — GOSPA, Missed Target Error, False Track Error, and Switching Error — the trends consistently indicated higher performance and lower error rates for the V2V-enabled priority track list as compared to the local fusion track list. The capability to significantly lower error rates in these metrics is crucial to marketing ADAS features in vehicles as both highly effective and safe for users. Being able to rely on V2V BSM data in complex design domains such as AIN provides a compelling case for further investment and development around cooperative perceptio6.

## VII. Acknowledgements

We would like to thank the CAVs sub-team at Virginia Tech's EcoCAR team for their continuous dedication as well as our faculty advisor Dr. Scott Huxtable. The work described was developed throughout Year 2 of the EcoCAR EV Challenge, sponsored by the U.S. Department of Energy (DOE), General Motors (GM), and MathWorks, managed by Argonne National Laboratory (ANL). We thank all our sponsors as well as Dr. Nick Goberville, the EcoCAR EV Challenge CAV organizer, for sharing their valuable resources and expertise.